\documentclass[journal,twoside,web]{ieeecolor}
\usepackage{generic}
\usepackage{cite}
\usepackage{amsmath,amssymb,amsfonts}
\usepackage{algorithmic}
\usepackage{graphicx}
\usepackage{pifont}
\usepackage{textcomp}
\usepackage{hyperref}       % hyperlinks
\usepackage{booktabs} 
\usepackage{subcaption}
\usepackage{epstopdf}
\def\BibTeX{{\rm B\kern-.05em{\sc i\kern-.025em b}\kern-.08em
    T\kern-.1667em\lower.7ex\hbox{E}\kern-.125emX}}
\begin{document}
\title{Synthesizing Mixed-type Electronic Health Records using Diffusion Models}
\author{Taha Ceritli, Ghadeer O.\ Ghosheh, Vinod Kumar Chauhan, Tingting Zhu, Andrew P.\ Creagh, and David~A.~Clifton
\thanks{All authors are affiliated with the Institute of Biomedical Engineering, Department of Engineering Science, University of Oxford,
Old Road Campus Research Building, Headington, Oxford OX3 7DQ (e-mail: {taha.ceritli,ghadeer.ghosheh,vinod.kumar,tingting.zhu,
andrew.creagh,david.clifton}@eng.ox.ac.uk).}}

\maketitle

\begin{abstract}
Electronic Health Records (EHRs) contain sensitive patient information, which presents privacy concerns when sharing such data. Synthetic data generation is a promising solution to mitigate these risks, often relying on deep generative models such as Generative Adversarial Networks (GANs). However, recent studies have shown that diffusion models offer several advantages over GANs, such as generation of more realistic synthetic data and stable training in generating various data modalities, including image, text, and sound. In this work, we investigate the potential of diffusion models for generating realistic mixed-type tabular EHRs, comparing TabDDPM model with existing methods on four datasets in terms of data quality, utility, privacy, and augmentation. Our experiments demonstrate that TabDDPM outperforms the state-of-the-art models across all evaluation metrics, except for privacy, which confirms the trade-off between privacy and utility.
\end{abstract}


\section{Introduction}
\label{sec:introduction}
% EHR and Privacy
\IEEEPARstart{T}{he} adoption of Electronic Health Records (EHRs) -- a rich source of patient information -- led to a rapid development of advanced machine learning models for timely diagnosis, treatment, and even efficient management of healthcare resources \cite{li2020behrt}. However, the sensitive nature of EHRs, including personal identifiers and confidential medical data, has led to strict laws worldwide to regulate data sharing practices, such as the General Data Protection Regulation (GDPR)\footnote{\url{https://gdpr-info.eu/} [Accessed on 25 February 2023]} in Europe and the Health Insurance Portability and Accountability Act (HIPAA)\footnote{\url{https://www.hhs.gov/hipaa/index.html} [Accessed on 25 February 2023]} in the United States. These restrictive measures limit the free access of EHRs by the research community, hindering the success of machine learning in healthcare \cite{gostin2009beyond}.

% De-identification 
Healthcare organizations and data owners often use the technique called \emph{de-identification} \cite{portability2012guidance} to mitigate privacy concerns of sharing EHRs. De-identification is the process of data anonymization via perturbing potentially identifiable patient attributes, e.g., through randomization, generalization or suppression. While de-identification is the most commonly used approach for sharing EHR \cite{johnson2016mimic}, the data could potentially be re-identified, e.g., using \emph{quasi-identifiers} which are pieces of information that are not of themselves unique identifiers, but when combined, become personally identifying information such as gender, birth dates, and postal codes \cite{erlich2014routes}.

% Synthetic data generation and deep generative models
An alternative technique to de-identification has been synthetically generating realistic patient records that do not belong to any real patient, yet, capture the characteristics of the original patient data \cite{choi2017generating,tucker2020generating,chen2021synthetic}. In the past years, deep generative models such as Variational AutoEncoders (VAEs) \cite{kingma2013auto} and Generative Adversarial Networks (GANs) \cite{goodfellow2014generative} showed great promise in generating realistic synthetic data by learning the underlying data distribution, across various data domains. In the case of EHR data, GANs are becoming an increasingly popular method where a wide range of models have been proposed. For instance, Choi et al.\ \cite{choi2017generating} proposed medGAN, a combination of an auto-encoder and a GAN to generate high dimensional multi-label discrete tabular EHR. Similarly, Torti and Fox \cite{Torfi2020CorGAN} proposed CorGAN to improve the quality of data by capturing local correlations in EHR. While GANs remain the state-of-the-art models for 
generating synthetic patient records, their main limitations lie in their unstable training process, which could result in mode collapse \cite{miyato2018spectral}.

% Diffusion models
Denoising Diffusion Probabilistic Models (DDPMs) are emerging as another approach for generating realistic synthetic data. DDPMs are a class of latent variable models that can learn the underlying distribution of data by transforming its samples into standard Gaussian noise and learning to denoise these corrupted samples back to their original forms. The trained model then allows us to generate new samples from the learned data distribution by applying this denoising process on standard Gaussian noise samples. In addition to their stable training, DDPMs have shown superior performance to GANs in generating image \cite{dhariwal2021diffusion}, audio \cite{kong2021diffwave,mittal2021symbolic}, and tabular data \cite{kotelnikov2022TabDDPM}. 

\begin{figure*}[htb!]
\begin{center}
    \includegraphics[width=\linewidth]{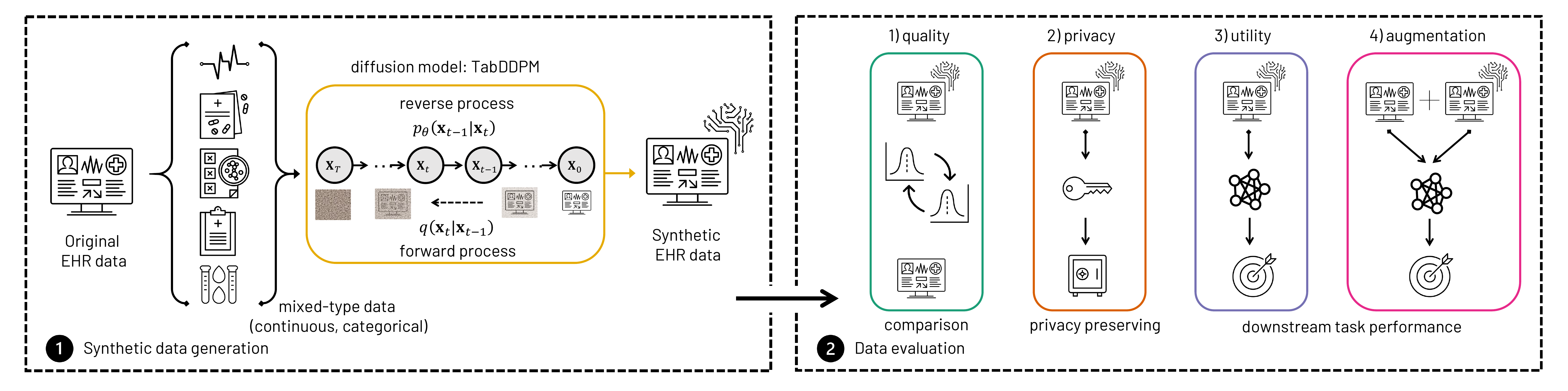}
    \caption{\textbf{An overview of our pipeline} that demonstrates the synthetic data generation process using TabDDPM (step 1) and how the generated data are evaluated based on quality, privacy, utility, and augmentation performance (step 2).}
    \label{fig:motivation-figure}
\end{center}    
\end{figure*}

Very recently, He et al.\ \cite{he2023meddiff} have proposed medDiff\footnote{Note that this work is a very recent work which has not been published in a peer-reviewed venue.} for EHRs demonstrating the benefits of diffusion models over existing methods. However, medDiff can generate only continuous values as it employs Gaussian diffusion process, assuming that the data is Gaussian distributed even though medical data often contains both continuous (e.g., blood test results) and categorical (e.g., sex and ethnicity) features. One can round the outputs of medDiff to obtain discrete values; however, a more principled approach would be to directly generate categorical data eliminating the need of this post-processing step and potentially leading to more accurate samples \cite{austin2021structured}.

In this paper, we use TabDDPM \cite{kotelnikov2022TabDDPM}, a type of DDPMs that can be applied to any tabular task. This model can work with mixed-type data as it combines Gaussian and Multinomial diffusion processes to respectively generate continuous and categorical data. The contributions of our work are summarized as follows:
\begin{itemize}
    \item We investigate the first application of diffusion models to generate mixed-type tabular EHRs with continuous and categorical features.
    \item We present a comprehensive comparison with the state-of-the-art generative models in terms of data quality, privacy, utility, and augmentation as shown in Figure \ref{fig:motivation-figure}.
    \item We show that our method achieves better performance than existing methods using four datasets in all applicable evaluation strategies except privacy which we draw attention for the first time in the context of EHR generation. 
\end{itemize}

The remainder of the paper is organized as follows. We first describe our methodology (Section~\ref{sec:methodology}) and then discuss the related works (Section~\ref{sec:related-work}). Finally, we present our experiments (Section~\ref{sec:experiments}) followed by a concluding remarks of our work and future plans (Section~\ref{sec:summary}).

\section{Methodology}
\label{sec:methodology}
This section provides a brief overview of diffusion models and then introduces TabDDPM.

\subsection{Diffusion Models}
\subsubsection*{Gaussian diffusion process}
Diffusion models \cite{dhariwal2021diffusion} are likelihood-based generative models that learn the distribution of a given dataset using two Markovian processes. The first process known as forward process, which is fixed, gradually adds noise to the samples until their distribution becomes an isotropic Gaussian. The other process called backward process denoises the corrupted samples allowing us to generate samples from the learned distribution.

Denoting our data distribution by $x_0 \sim q(x_0)$, we define a Markovian process $q$ that produces the corrupted samples $x_{1:T}$ by gradually adding noise to the data at each time step $t$ as follows:
\begin{eqnarray}
    q(x_t|x_{t-1}) &=& \mathcal{N}(x_t; \sqrt{1- \beta_t} x_{t-1}, \beta_t I), 
\end{eqnarray}
where $\mathcal{N}$ denotes Gaussian distribution and $\beta_t \in (0, 1)$ is called the variance schedule. 

Notice that this transition process $q$ scales the data and adds Gaussian noise to it at each transition. As the process is Markovian, we obtain $q(x_t|x_0) = \mathcal{N}(x_t; \sqrt{1- \beta_t}^t x_0, \beta_t^t I)$ when using fixed variance schedule. Therefore, the distribution $q(x_T)$ becomes nearly an isotropic Gaussian distribution, i.e., $\mathcal{N}(0, 1)$, given sufficiently large $T$ \cite{nichol2021improved}. As a result, when $q(x_{t-1} | x_t)$ is known, we can generate new samples from $q(x_0)$ starting with a Gaussian noise. However, as we cannot easily estimate $q(x_{t-1} | x_t)$, the typical approach has been training a neural network $p_\theta(x_{t-1} | x_t)$ to approximate these conditional probabilities defined as follows:
\begin{eqnarray}
p_\theta(x_{t-1}|x_t) &=& \mathcal{N}(x_{t-1}; \mu_\theta(x_t, t), \Sigma_\theta(x_t, t)), 
\end{eqnarray}
where the mean $\mu_\theta$ and covariance $\Sigma_\theta$ are parametrized using a deep neural network. To train this neural network so that $p_\theta(x_0)$ learns the true data distribution $q(x_0)$, we maximize the data log-likelihood as follows:

\begin{align}
\log p_\theta(x_0) &= \log \int p(x_{0:T}) dx_{1:T}, \nonumber\\
&= \log \int \frac{p(x_{0:T})q(x_{1:T}|x_0)}{q(x_{1:T}|x_0)} dx_{1:T}, \nonumber\\
&\geq \mathbb{E}_{q(x_{1:T}|x_0)}\left[\log \frac{p(x_{0:T})}{q(x_{1:T}|x_0)}\right] dx_{1:T} \nonumber\\, \\
&= \underbrace{\mathbb{E}_{q(x_1|x_0)}[\log p_\theta(x_0|x_1)]}_{L_0} \nonumber\\ 
& \hspace{0.5em} - \sum_{t=2}^{T} \underbrace{\mathbb{E}_{q(x_t|x_0)} [KL(q(x_{t-1}|x_t, x_0) \parallel p_\theta (x_{t-1}|x_t))}_{L_{t-1}}] \nonumber\\
& \hspace{0.5em} - \underbrace{KL(q(x_T |x_0) \parallel p(x_T))}_{L_T},  \label{eq:elbo} 
\end{align}
where KL denotes Kullback-Liebler (KL) divergence. Here, $L_0$ can be interpreted as a reconstruction term predicting the log probability of the original sample given the first-step latent. $L_t$ measures how close the neural network transition $p_\theta (x_{t-1}|x_t)$ approximates to the ground-truth transition $q(x_{t-1}|x_t, x_0)$. Finally,  $L_T$ measures how close the final corrupted sample to the standard Gaussian prior. Note that we apply Jensen's Inequality at the step 3 of Equation \ref{eq:elbo}.

Ho et al.\ \cite{ho2020denoising} reparametrize $L_t$ in Equation \ref{eq:elbo} to predict the Gaussian noise $\epsilon_t$ rather than $\mu(x_t, t)$. The authors further show that training the diffusion model works better with the resulting simplified objective function:
\begin{eqnarray}
 L_t^{\text{simple}} &=& \mathbb{E}_{x_0, t, \epsilon} || \epsilon - \epsilon_{\theta}(x_t, t) ||^2,
\end{eqnarray}
where $\epsilon$ denotes the noise component for the noisy data sample $x_t$. In practice, this can be simplified to the sum of mean-squared errors between $\epsilon$ and $\epsilon_{\theta}(x_t, t)$ over all timesteps.

\subsubsection*{Multinomial diffusion process}
Gaussian diffusion process assumes that the data $x_0$ is Gaussian distributed. Hoogeboom et al.\ \cite{hoogeboom2021argmax} adapt diffusion models to categorical data by proposing the following Markovian process $q$:
\begin{eqnarray}
    q(x_t|x_{t-1}) &= \mathcal{C}at(x_t; (1- \beta_t) x_{t-1} + \beta_t/K), 
\end{eqnarray}
where $\mathcal{C}at$ denotes Categorical distribution. This process provides a $\beta_t$ chance of resampling a category uniformly at each transition. As the process is Markovian, the distribution $q(x_T)$ becomes nearly a uniform distribution over the categories for large $T$. Therefore, we can employ the reverse process to generate new samples of a categorical feature starting with uniform distribution over its categories.  

Note that we define a new forward process for multinomial diffusion process; however, our objective function stays the same which is given in Equation \ref{eq:elbo}.

\subsection{TabDDPM}
Kotelnikov et al.\ \cite{kotelnikov2022TabDDPM} propose TabDDPM for classification and regression tasks using tabular data. The reverse process is built using a multilayer perceptron (MLP) with the following architecture:
\begin{eqnarray}
t_{emb} &=& \text{Linear}(\text{SiLU}(\text{Linear}(\text{SinTimeEmb}(t)))), \nonumber\\
y_{emb} &=& \text{Embedding}(y),\nonumber\\
x &=& \text{Linear}(x_{in}) + t_{emb},
\end{eqnarray}
where $x_{\text{in}}$, $y$, and $t$ respectively denote a data input, its target variable, and a timestep. Here, \text{SinTimeEmb} refers to a sinusoidal time embedding as in \cite{dhariwal2021diffusion,nichol2021improved,kotelnikov2022TabDDPM}, while \text{SiLU} denotes the sigmoid linear unit function.

MLP is trained by minimizing a sum of mean-squared error $L_{\text{simple}}$ for the continuous features and the KL divergences $L^i_t$ for each categorical feature. The combined objective function becomes as follows:
 \begin{eqnarray}
L_t^{\text{TabDDPM}} = L_t^{\text{simple}} + \frac{\sum_i L_t^{i}}{C},
\end{eqnarray}
where $C$ denotes the number of categorical features.

TabDDPM adapts the reverse process to the categorical data by applying the softmax function to the outputs of MLP. Additionally, it learns a class-conditional model for classification tasks, i.e., $p_\theta(x_{t-1}|x_t, y)$.

Note that we omit using a target variable in the reverse process when the data of interest does not contain a target variable. In such cases, we learn the joint distribution of the features themselves.

\section{Related Works}
\label{sec:related-work}
We divide the related works into two categories: (i) existing methods for generating synthetic EHR data and (ii) existing diffusion models used for synthetic data generation.

\subsection{Synthetic Data Generation for EHR}
\label{subsec_synthetic}

The most common approach used for generating EHRs has been using GANs, which consist of two neural networks named generator and discriminator~\cite{goodfellow2014generative}. These two networks are simultaneously trained such that the generator learns to generate plausible samples, while the discriminator learns to distinguish these realistic fake samples from the real samples. 

One of the challenges for generating EHRs using GANs has been the mode collapse problem. medGAN \cite{choi2017generating}, which is one of the early adoptions of GANs for EHRs, addresses this issue by introducing minibatch-averaging. This model has been further modified and advanced by many researchers \cite{baowaly2019synthesizing,Torfi2020CorGAN,tantipongpipat2021differentially}. Ghosheh et al.\ \cite{ghosheh2022review} provide a comprehensive overview of the existing approaches based on GANs to generate EHRs. These methods tackle different data types (e.g., tabular or time-series), different variable types (e.g., continuous, categorical or mixed) and various applications (e.g., privacy-preservation and treatment effects estimation).

\subsection{Diffusion Models}
\label{subsec_diff_literature}

Following the contributions presented by \cite{song2020improved} and \cite{ho2020denoising}, several new advancements have been proposed to improve the capabilities of diffusion models. For example,  \cite{nichol2021improved} proposed cosine-based variance schedule, introduced a neural network for learning reverse process variances and proposed a hybrid objective function. \cite{ho2021} generated high-quality images using a pipeline of multiple diffusion models with increasing resolutions. Other recent notable extensions of diffusion models include \cite{saharia2022photorealistic,ramesh2022hierarchical}. 

After the superior performance achieved on ImageNet~\cite{dhariwal2021diffusion}, diffusion models were used for a wide variety of applications such as music generation \cite{mittal2021symbolic} and audio generation \cite{kong2021diffwave}. Recently, Kotelnikov et al.\ \cite{kotelnikov2022TabDDPM} develop a diffusion model called TabDDPM to generate tabular data from various domains. Their model can generate both a number of features and a target variable which are then used for a classification or regression task. We note that TabDDPM has not been studied in the context of EHR generation. 

He et al.\ \cite{he2023meddiff} propose the most related work to ours. Their proposed model called medDiff can generate high-quality numerical EHR data using Gaussian diffusion process. medDiff is further equipped with a numerical method to accelerate its data generation process. However, the authors do not consider the privacy aspect of diffusion models for EHR generation and does not use Multinomial diffusion process for generating categorical features.

\section{Experiments}
\label{sec:experiments}
In this section, we initially describe our experimental setup (Section~\ref{sec:exp-setup}) and present our findings (Section~\ref{sec:exp-results}). The goal of our experiments is to evaluate the models in terms of data quality, privacy, utility and augmentation. Note that we only report the data quality and privacy metrics for the MIMIC-III dataset as it does not include any target variable, following the preprocessing of previous works \cite{choi2017generating,Torfi2020CorGAN}. Upon the acceptance of the paper, we are going to release
our code that reproduces our experiments.

\subsection{Experimental Setup}
\label{sec:exp-setup}
Below we describe our datasets, evaluation metrics and baseline models.
\subsubsection*{Data} 
We conduct experiments using four datasets:
\begin{itemize}
    \item MIMIC-III \cite{johnson2016mimic}: A publicly available dataset consisting of medical records of more than 46,000 patients. Following the previous works \cite{choi2017generating,Torfi2020CorGAN}, we extracted the World Health Organization’s Ninth Revision, International Classification of Diseases (ICD-9) codes\footnote{\url{https://www.cdc.gov/nchs/icd/icd9cm.htm} [Accessed on 6 October 2022]}, grouped them by generalizing up to their first 3 digits and aggregated a patient's longitudinal record over time to obtain a single fixed-size vector of length 1,071, where the entries are converted into binary values.
    
    \item Pima Indians Diabetes\footnote{available at \url{https://www.kaggle.com/datasets/uciml/pima-indians-diabetes-database [Accessed on 16 February 2023]}}: A set of medical measurements such as blood pressure and body mass index collected from 768 patients. Discarding the age column, we obtain seven continuous features which is then used to predict whether a patient has diabetes or not. Here, our goal is to evaluate the capabilities of the models in dealing with only continuous features. 
    
    \item Indian Liver Patient Dataset (ILPD) \footnote{available at \url{https://archive.ics.uci.edu/ml/datasets/ILPD+\%28Indian+Liver+Patient+Dataset\%29 [Accessed on 20 February 2023]}}: A collection of medical information which is used to predicted whether a person is a liver patient or not. This dataset contains 416 liver patient records and 167 non-liver patient records. Discarding the age and sex column, we obtain eight continuous features. Similarly to the Pima dataset, we discard the categorical features to compare the models in terms of dealing with continuous features only.
    
    \item Stroke prediction\footnote{available at \url{https://www.kaggle.com/datasets/fedesoriano/stroke-prediction-dataset} [Accessed on 17 February 2023]}: A dataset used to predict whether a patient is likely to get stroke based on the features such as gender, age, various diseases, and smoking status. The dataset contains 5,110 observations where each observation consists of 2 continuous and 8 categorical features.
\end{itemize}

We normalize numerical columns using min-max normalizer so that the data values are in the range of $[0.0, 1.0]$. The datasets are split with the ratio of 70\%, 15\% and 15\% respectively for the training, validation and test sets. Note that the ILPD and Pima are relatively small-scale datasets which are chosen to explore the effectiveness of the models in dealing with in-sufficient training data. 

\begin{table}[htb!]
\centering
\normalsize
\caption{\textbf{Summary of our experimental setup.} The table shows our datasets and the evaluation aspects they are considered for. Here, \ding{52} denotes whether a dataset is used for a particular evaluation aspect. Note that we use the MIMIC-III dataset only for the data quality and privacy evaluations as it does not include any target variable, following the previous works \cite{choi2017generating,Torfi2020CorGAN}.}
\resizebox{.5\textwidth}{!}{%
\begin{tabular}{llcccc}
\toprule
Name & Data type & Quality &  Privacy & Utiliy & Augmentation\\
\midrule
MIMIC-III & Binary & \ding{52} & \ding{52} &  & \\
Pima & Continuous & \ding{52} & \ding{52} & \ding{52} & \ding{52}\\
ILPD & Continuous & \ding{52} & \ding{52} & \ding{52} & \ding{52}\\
Stroke & Mixed & \ding{52}  & \ding{52} & \ding{52} & \ding{52}\\
\bottomrule 
\end{tabular}
}
\label{tab:datasets}
\end{table}

\subsubsection*{Evaluation} 
We evaluate the models using the following metrics:
\begin{itemize}
    \item \textbf{Data quality}:
    \begin{itemize}
    \item Dimension-wise probability: To check whether a model correctly learns the distribution of the real data columns, we report the probability for each dimension, which is a Bernoulli success probability for a binary column and a Gaussian mean parameter for a numerical column.
    
    \item Dimension-wise prediction: To evaluate how well a model captures the inter-dimensional relationships of the real data, we report the predictive performance of a binary classifier (i.e., logistic regression classifier), where one dimension is chosen as the target variable and the remaining dimensions are used as the features. In particular, we use F1-score with the threshold of 0.5 following the related works.

    \item Maximum mean discrepancy (MMD) \cite{gretton2012kernel}: To analyze to what extend synthetic data preserves the joint distribution of the real data, we calculate MMD following the works of \cite{ghosheh2022review,li2021generating}. Lower MMD values mean that the synthetic data captures the joint distribution of the real data better.
    \end{itemize}
    
    \item \textbf{Privacy}: To quantify how much a synthetically generated dataset reveals the sensitive content in a real dataset, we use the median Distance to Closest Record (DCR) \cite{kotelnikov2022TabDDPM, zhao2021ctab} and Membership Inference Risk (MIR) \cite{yan2022multifaceted} which are calculated as follows:
    \begin{itemize}
    \item DCR: We calculate the minimum distance of each generated sample to real samples and take the median of these distances as DCR values. Low DCR values indicate that the synthetically generated dataset violates the privacy requirements as its samples are similar to the real samples.
    \item MIR: We calculate the Euclidean distance between each synthetic sample and each real sample. Given a distance threshold, we claim that a sample is in the real training dataset if there exists at least one synthetic sample with a distance smaller than the threshold. We then calculate MIR as the F1 score. Low MIR is better in terms of privacy as it means that the membership status of a sample cannot be identified.
    \end{itemize}
    
    \item \textbf{Data utility}: To evaluate the utility of the synthetically generated data for a downstream task, we consider a binary classification problem as in \cite{Torfi2020CorGAN}. We train two sets of seven predictive models (namely decision tree, random forest, logistic regression, multi-layer perceptron, gradient boosting classifier, k-nearest neighbor and XGBoost) using real and synthetic data. The trained predictive models are then evaluated on the same real test data. We calculate overall accuracy, area under the receiver operating characteristics curve (AUROC), area under the precision recall curve (AUPRC) and F1-Score, and report their average across the predictive models. Finally, we conclude that synthetic data has a good utility if the predictive performance of these models are similar. 
    
    \item \textbf{Data augmentation}: To investigate whether synthetically generated data can be beneficial for performance in the case of in-sufficient and potentially imbalanced training data, we augment real training data with varying amounts of synthetically generated data. We generate equal number of synthetic samples per class which helps in handling data-imbalance issues. The resulting dataset is then used to train classifiers which we evaluate in terms of F1-score on the test set. Note that we train the same seven classifiers used in data utility evaluations and calculate their average. 
\end{itemize}

Note that once a model is trained, we generate 5 sets of synthetic data using 5 random seeds, each of which is evaluated using 10 random seeds.

\begin{figure*}[h!]
\centering
\begin{subfigure}{\linewidth}
    \centering
    \includegraphics[width=.9\textwidth]{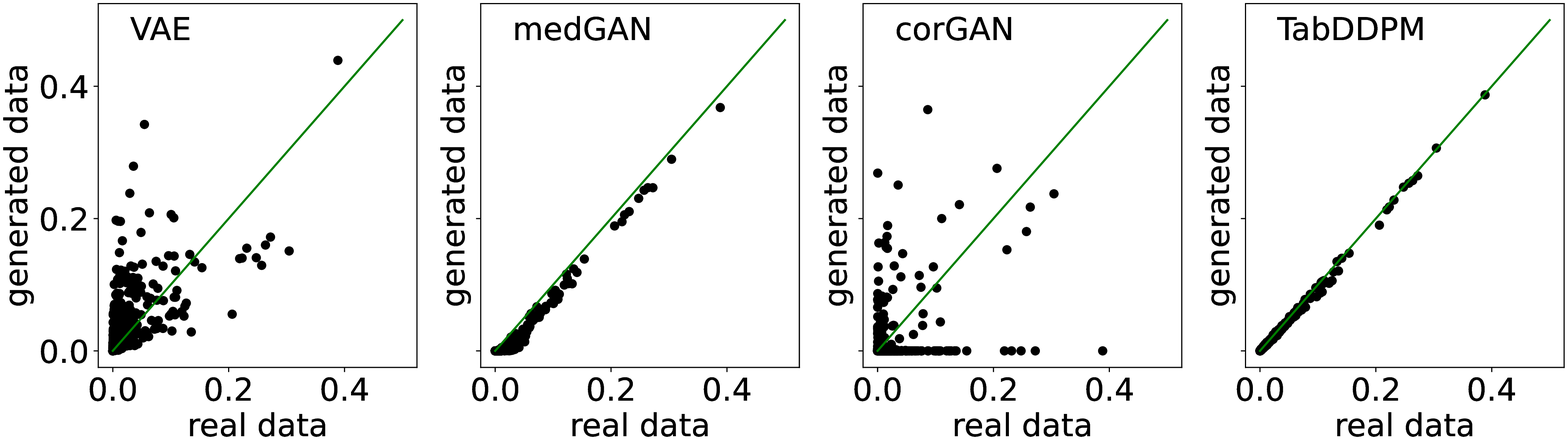}
    \caption{Scatter plot of dimension-wise probability.}
    \label{fig:dim-wise-probability}
\end{subfigure}
\begin{subfigure}{\linewidth}
    \centering
    \includegraphics[width=.9\textwidth]{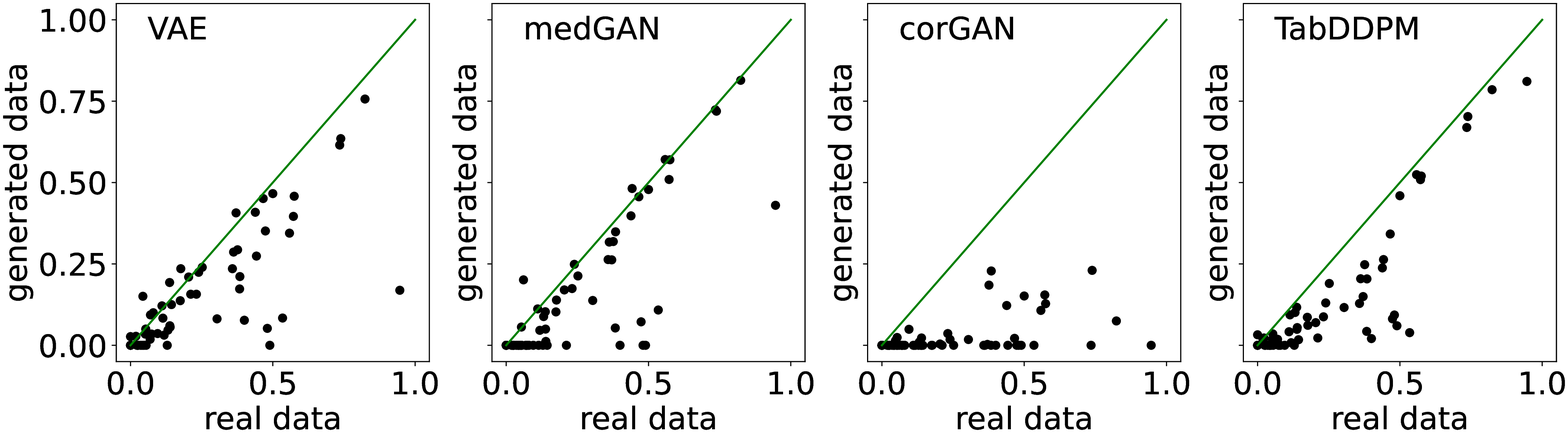}
    \caption{Scatter plot of dimension-wise prediction.}
    \label{fig:dim-wise-prediction}
\end{subfigure}
\caption{\textbf{Data quality evaluations using the MIMIC-III dataset illustrating the improved performance of TabDDPM over the competitor models.} Each dot represents one of 1,017 ICD-9 codes. The diagonal line indicates the ideal performance where the real and synthetic data show identical quality. In a), the x-axis represents the Bernoulli success probability/Gaussian mean parameter for the real dataset, and y-axis the probability for the synthetic counterpart generated by each model. In b), the x-axis represents the F1-score of the logistic regression classifier trained on the real dataset, while the y-axis represents the F1-score of the classifier trained on the synthetic counterpart generated by each model.}
\end{figure*}

\begin{table*}[htb!]
\centering
\normalsize
\caption{\textbf{Performance of the models in terms of data quality}. The RMSE values of dimension-wise probability and dimension-wise prediction, calculated using the real data and the synthetically generated data, as well as MMD values.}
\resizebox{\textwidth}{!}{%
\begin{tabular}{@{\extracolsep{2pt}}lcccccccccccc@{}}
\toprule
& \multicolumn{3}{c}{MIMIC-III} & \multicolumn{3}{c}{ILPD} & \multicolumn{3}{c}{Pima} & \multicolumn{3}{c}{Stroke} \\
\cline{2-4} \cline{5-7} \cline{8-10} \cline{11-13}\\
& DWProb$\downarrow$ & DWPred$\downarrow$ & MMD$\downarrow$ & DWProb$\downarrow$ &  DWPred$\downarrow$ & MMD$\downarrow$ & DWProb$\downarrow$ &  DWPred$\downarrow$ & MMD$\downarrow$ & DWProb$\downarrow$ &  DWPred$\downarrow$ & MMD$\downarrow$\\
\midrule
VAE & 0.9e-3 & 0.0286 & 0.0021 & 1.9e-1 & 1.8e+1 & 1.262  & 1.7e-2 & 1.7e-1 & 0.184 & 1.4e-2 & 0.1e-2 & 0.0819\\
medGAN & 7.5e-5 & 0.0265 & 0.0111 & 1.0e-2 & 8.3e-3 & 0.063 & 0.7e-3 & 1.1e-2 & 0.012 & 9.1e-5 & 1.0e-2 & 0.0064\\
corGAN & 1.2e-3 & 0.0943 & 0.0296 & 1.5e-1 & 3.5e-2  & 0.805 & 9.1e-2 & 0.9e-2 & 0.516 & 3.1e-2 & 2.5e-2 & 0.2625\\
TabDDPM & \textbf{0.4e-5} & \textbf{0.0254} & \textbf{0.0002} & \textbf{7.7e-5} & \textbf{3.5e-6} & \textbf{0.001} & \textbf{0.3e-4} & \textbf{0.1e-4} & \textbf{0.001} & \textbf{8.2e-5} & \textbf{1.2e-6} & \textbf{0.0007} \\
\bottomrule 
\end{tabular}
}
\label{tab:data-quality}
\end{table*}

\subsubsection*{Baseline Methods} 
We compare our proposed model with the three baseline methods described below. While medGAN and corGAN are chosen as the early adoptions of GANs for generating EHRs, we include VAEs in our evaluations to make a comparison between the three types of models.
\begin{itemize}
    \item \textbf{Variational Autoencoder (VAE)}: We use VAEs \cite{kingma2013auto} with the encoder and the decoder built using fully connected neural networks following \cite{choi2017generating}. Each component has two hidden layers with the size of 128. 
    \item \textbf{medGAN}: A GAN-based model adapted to generate discrete tabular EHRs, by incorporating an additional auto-encoder and minibatch-averaging to cope with the mode collapse problem~\cite{choi2017generating}. The discriminator of medGAN is a fully connected network with 3 layers where we use ReLU as the activation function for the layers except the last layer for which we use the sigmoid function. Similarly, we use a fully connected network with 2 layers for the generator where we use ReLU for the first layer and the tanh function for the second layer.
    \item \textbf{corGAN}: A GAN-based model that applies 1-dimensional Convolutional Autoencoder, to capture the feature correlations in tabular datasets~\cite{Torfi2020CorGAN}. The auto-encoder, discriminator and generator of corGAN are built using convolutional layers. The number of layers varies between 3 and 6 which is adjusted for the number of features in the data. We apply batch normalization, LeakyReLU, ReLU, the tahh and the sigmoid functions.
\end{itemize}

We used ADAM optimizer for 500 epochs with the batch size of 256 to train the models. Additionally, we pretrain auto-encoders of medGAN and corGAN for 100 iterations.

\subsection{Experimental Results}
\label{sec:exp-results}
We discuss our results on data quality, privacy, utility, and augmentation below.
\subsubsection*{Data quality}
We observed a notable improvement in performance when using TabDDPM compared to VAE, medGAN, and corGAN in terms of dimension-wise probability, as illustrated for the MIMIC-III dataset in Fig.\ \ref{fig:dim-wise-probability}. Additionally, Table \ref{tab:data-utility} shows that TabDDPM consistently generates more realistic samples than the baseline models across all datasets, as evidenced by the lower root mean square error (RMSE) values.

Fig.\ \ref{fig:dim-wise-prediction} demonstrates the models' performance in terms of dimension-wise prediction. As shown by the RMSE values in Table \ref{tab:data-quality}, TabDDPM outperforms the baseline models in this aspect, too. These results confirm the superior capability of diffusion models to VAE and GANs in modeling the inter-dimensional relationships of the real data. Similarly, a similar trend in terms of MMD values indicates that TabDDPM preserves the joint distribution of the real data better than the competitor methods.

\begin{table*}[htb!]
\centering
\normalsize
\caption{\textbf{Performance of the models in terms of data utility}. The Accuracy, F1-Score, AUPRC and AUROC for the Pima, ILPD and Stroke Prediction datasets.}
% \begin{tabular}{lcccccccc}
\resizebox{\textwidth}{!}{%
\begin{tabular}{@{\extracolsep{2pt}}lcccccccccccc@{}}
\toprule
& \multicolumn{4}{c}{Pima} & \multicolumn{4}{c}{ILPD} & \multicolumn{4}{c}{Stroke}\\
\cline{2-5} \cline{6-9} \cline{10-13}\\
& Accuracy & F1-Score &  AUROC & AUPRC & Accuracy & F1-Score &  AUROC & AUPRC & Accuracy & F1-Score &  AUROC & AUPRC\\
\midrule
VAE & 0.38 $\pm$ 0.01 & 0.30 $\pm$ 0.03 & 0.57 $\pm$ 0.07 & 0.45 $\pm$ 0.08 & 0.51 $\pm$ 0.01 & 0.34 $\pm$ 0.02 & 0.49 $\pm$ 0.02 & 0.29 $\pm$ 0.03 & 0.60 $\pm$ 0.11 & 0.39 $\pm$ 0.05 & 0.46 $\pm$ 0.01 & 0.05 $\pm$ 0.00\\
medGAN & 0.67 $\pm$ 0.03 & 0.64 $\pm$ 0.03 & 0.72 $\pm$ 0.05 & 0.60 $\pm$ 0.07 & \textbf{0.73 $\pm$ 0.01} & 0.51 $\pm$ 0.05 & 0.57 $\pm$ 0.05 & 0.38 $\pm$ 0.06 & \textbf{0.90} $\pm$ 0.01 & \textbf{0.57 $\pm$ 0.02} & 0.68 $\pm$ 0.09 & 0.13 $\pm$ 0.04 \\
corGAN & 0.53 $\pm$ 0.02 & 0.45 $\pm$ 0.02 & 0.51 $\pm$ 0.03 & 0.45 $\pm$ 0.03 & 0.59 $\pm$ 0.08 & 0.40 $\pm$ 0.02 & 0.44 $\pm$ 0.04 & 0.26 $\pm$ 0.02 & 0.71 $\pm$ 0.09 & 0.43 $\pm$ 0.04 & 0.46 $\pm$ 0.02 & 0.06 $\pm$ 0.00 \\
TabDDPM & \textbf{0.68 $\pm$ 0.02} & \textbf{0.69 $\pm$ 0.02} & \textbf{0.79 $\pm$ 0.04} & \textbf{0.68 $\pm$ 0.07} & 0.66 $\pm$ 0.03 & \textbf{0.63 $\pm$ 0.03} & \textbf{0.73 $\pm$ 0.03} & \textbf{0.44 $\pm$ 0.03} & 0.68 $\pm$ 0.02 & 0.50 $\pm$ 0.01 & \textbf{0.78 $\pm$ 0.04} & \textbf{0.18 $\pm$ 0.05} \\
\hline
Real Data & 0.70 $\pm$ 0.02 & 0.69 $\pm$ 0.02 & 0.79 $\pm$ 0.03 & 0.68 $\pm$ 0.04 & 0.65 $\pm$ 0.04 & 0.63 $\pm$ 0.03 & 0.73 $\pm$ 0.03 & 0.46 $\pm$ 0.04 & 0.68 $\pm$ 0.02 & 0.50 $\pm$ 0.02 & 0.81 $\pm$ 0.03 & 0.19 $\pm$ 0.05 \\
\bottomrule
\end{tabular}
}
\label{tab:data-utility}
\end{table*}

\begin{table}[htb!]
\centering
\normalsize
\caption{\textbf{Performance of the models in terms of privacy}. The DCR and MIR values calculated using real and synthetically generated data.}
\resizebox{.5\textwidth}{!}{%
\begin{tabular}{@{\extracolsep{4pt}}lcccccccc@{}}
\toprule
& \multicolumn{2}{c}{MIMIC-III} & \multicolumn{2}{c}{ILPD} & \multicolumn{2}{c}{Pima} & \multicolumn{2}{c}{Stroke} \\
\cline{2-3} \cline{4-5} \cline{6-7} \cline{8-9}\\
& DCR$\uparrow$  & MIR$\downarrow$ & DCR$\uparrow$ & MIR$\downarrow$ &  DCR$\uparrow$  & MIR$\downarrow$ &  DCR$\uparrow$  & MIR$\downarrow$\\
\midrule
VAE & \textbf{4.0} & \textbf{0.127} & \textbf{1102} & \textbf{0.005} & 19 & 0.869 & \textbf{4.5} & \textbf{0.001} \\
medGAN & 1.4 & 0.408 & 25 & 0.884  & 21 & 0.899 & 2.1 & 0.902\\
corGAN & 2.2 & 0.237 & 1098 & 0.113  & \textbf{36} & \textbf{0.691} & 2.3 & 0.711 \\
TabDDPM & 2.4 & 0.291 & 7 & 0.903 & 6 & 0.902 & 1.6 & 0.898\\
\bottomrule 
\end{tabular}
}
\label{tab:privacy}
\end{table}

\begin{figure*}[htb!]
\begin{center}
    \includegraphics[width=.9\linewidth]{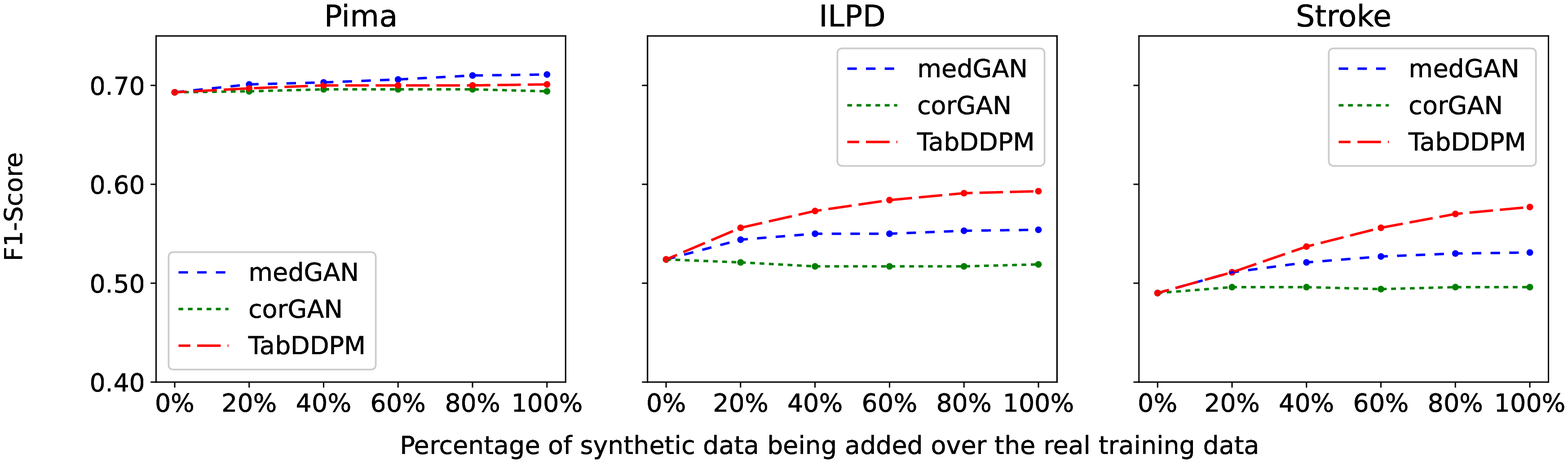}
    \caption{\textbf{Performance gain obtained using synthetic data via data augmentation} We use all available real training data and vary the amounts of synthetic data added. Therefore, $0\%$ denotes the performance obtained without data augmentation while $100\%$ denotes the performance obtained when augmenting the real data with the same number of synthetic samples.}
    \label{fig:data-augmentation}
\end{center}    
\end{figure*}
\subsubsection*{Privacy}
In terms of protecting the privacy of real data, Table \ref{tab:privacy} shows that TabDDPM is generally less effective than its competitor methods. It is important to note that DCR and MIR rely on distances between generated and real samples, which means that TabDDPM is capable of generating more realistic synthetic samples than alternative models. However, this increased realism also raises potential privacy risks associated with the original data. This trade-off between quality and privacy is discussed in earlier works based on GANs \cite{yan2022multifaceted,ghosheh2022review}. However, this study draws attention to the privacy concerns associated with EHR generation using diffusion models for the first time, which have been overlooked in previous studies.

\subsubsection*{Data Utility}
Table \ref{tab:data-utility} displays the performance of the models on a binary classification task across three datasets. We consistently observe that TabDDPM outperforms the competitor models across various metrics for all datasets. Furthermore, the results obtained using TabDDPM are comparable to those obtained using real data. Note that medGAN is better than TabDDPM in terms of the accuracy and F1-Score metrics for the Stroke Prediction dataset; however, these metrics can be misleading as this dataset is highly imbalanced and that we may want to give equal importance to each class. Additionally, VAE, medGAN and corGAN are not directly applicable to this dataset as it has non-binary categorical features; however, we have adapted them by applying ordinal encoding and minmax normalizer to the categorical features.

\subsubsection*{Data augmentation}

Figure \ref{fig:data-augmentation} depicts the impact of data augmentation on classifier performance. The figure shows that augmenting real training data with synthetic data generally improves performance, with TabDDPM consistently improving performance across three datasets. The performance gain is relatively small for the Pima dataset, which does not suffer from major data imbalance issues. However, the gap is larger for ILPD, with TabDDPM leading to the highest performance gain. Note that TabDDPM allows for easy generation of equal numbers of samples by modeling the distribution of the target variable, whereas this is not directly possible for VAE, medGAN, and corGAN, as the samples they produce can greatly favor one class (either majority or minority). In fact, VAE produced samples exclusively from the majority or minority class, preventing us from evaluating its performance in data augmentation experiments. For medGAN and corGAN, we needed to generate high numbers of synthetic samples and downsample them so that we have equal number of samples per class.

Recall that the Pima and ILPD datasets are relatively small-scale datasets with low numbers of training samples. Our results indicate that TabDDPM can deal with even such datasets, improving the performance of predictive models.

\section{Conclusions \& Future Work}
\label{sec:summary}
In this paper, we have explored the effectiveness of TabDDPM for synthetically generating realistic tabular electronic health records (EHRs). To the best of our knowledge, this is the first attempt to generate tabular EHRs with continuous and categorical features using diffusion models. We conducted a comprehensive evaluation of TabDDPM against state-of-the-art generative models across four datasets, analyzing their performance in terms of data quality, utility, privacy, and augmentation. Our results demonstrate that TabDDPM outperforms other models in terms of all evaluation metrics, except for privacy. This confirms the trade-off between privacy and utility in EHR generation, which has been largely overlooked in previous diffusion model-based studies. In future work, we plan to address this issue and explore ways to enhance the privacy of diffusion models in order to achieve a better balance between privacy and utility.

Synthetically generating realistic EHRs data has an essential role in scaling machine learning works for various healthcare applications, including clinical decision support systems and clinical research applications. The robustness of DDPMs could provide a lucrative option for disseminating deep generative models, addressing a gap currently faced by many GAN-based models. Furthermore, this work could be extended to address data-related challenges in healthcare research, beyond the lack of open-access datasets. For instance, similarly to many of the GAN and VAE works, the diffusion models could be adapted to handle irregularly sampled data and the presence of missing data. We hope that the promising results in this work motivate future research directions using diffusion models in building robust deep generative models for healthcare applications without compromising the quality of the generated synthetic data. 

\bibliographystyle{ieeetr}
\bibliography{references}

\end{document}